\documentclass[11pt]{article}

\usepackage[preprint]{acl}

\usepackage[main=british]{babel}
\useshorthands*{"}
\defineshorthand{-}{\babelhyphen{hard}}
\defineshorthand{"=}{\babelhyphen{–}}

\usepackage{times}
\usepackage{latexsym}
\usepackage[T1]{fontenc}
\usepackage[utf8]{inputenc}
\usepackage{microtype}
\usepackage{inconsolata}
\usepackage{graphicx}
\usepackage{booktabs}
\usepackage{enumitem}
\usepackage{makecell}
\usepackage{multirow}

\defcitealias{iptcMediaTopics}{IPTC, 2026}

\title{Reading the News: Adapting Large Language Models to\\Swedish Journalism Through Continued Pre-Training}

\author{
 \textbf{Lukas Borggren\textsuperscript{1,2}},
 \textbf{Jenny Kunz\textsuperscript{1}},
 \textbf{Marco Kuhlmann\textsuperscript{1}}
\\
\\
 \textsuperscript{1}Linköping University
 \\
 \textsuperscript{2}Bonnier News
\\
 \texttt{firstname.lastname@liu.se}
}

\begin{document}
\maketitle
\begin{abstract}
Large language models are increasingly capable in general, but their utility can remain modest in niche or understudied areas. One approach to address this limitation is to specialise existing models through additional training on target-domain corpora. In this work, we investigate such continued pre-training for adapting large language models to Swedish journalism, using a high-quality dataset that we curate from millions of news articles. To evaluate the adaptation efficacy, we also construct a novel domain-specific benchmark that covers six editorial tasks. Through full and parameter-efficient fine-tuning across two model sizes, we find that continued pre-training yields benefits in the target domain, but only when paired with experience replay to mitigate forgetting. We observe consistent enhancements in the models' generation quality and factual knowledge, but not their proficiency in discriminative tasks. Exploring a training-free method to facilitate instruction following, we see further improvements, but exclusively for models trained with low-rank adaptation. Crucially, we demonstrate the importance of targeted evaluation in the adaptation process, as an existing Swedish benchmark largely fails to capture the models' in-domain performance gains.
\end{abstract}

\section{Introduction}
\label{sec:intro}
News text has historically been near-ubiquitous in natural language processing via resources like the WSJ Penn Treebank \citep{marcusBuildingLargeAnnotated1993}, Reuters corpora \citep{lewisRCV1NewBenchmark2004}, and CNN/Daily~Mail \citep{hermannTeachingMachinesRead2015}. In the current paradigm of generative large language models (LLMs), complex problems like fake news detection \citep{tongGenerateFirstThen2025} and personalised headline generation \citep{wanPersonalizingNewsHeadlines2026} are studied. However, journalism has rarely been treated as a target domain for LLM adaptation; to the best of our knowledge, merely one study has investigated it \citep{yaoNewsGPTLarge2024}. Other work has created foundational LLMs for journalism, but focuses on pre-training from scratch for high-resource languages \citep{wangMediaGPTLargeLanguage2023, wuBloombergGPTLargeLanguage2023}. Such training demands vast amounts of compute and data, posing a barrier to model development for lower-resource languages.

Continued pre-training (CPT) on specialised corpora is commonly used to adapt general-purpose LLMs to domains like finance \citep{liCFGPTChineseFinancial2023}, law \citep{colomboSaulLM54BSaulLM141BScaling2024}, and biomedicine \citep{labrakBioMistralCollectionOpenSource2024}. It is similarly effective for improving models on resource-constrained languages, including the Scandinavian languages Danish \cite{zhangSnakModelLessonsLearned2025}, Norwegian \citep{samuelSmallLanguagesBig2025}, Icelandic \citep{gogoulouContinualLearningLanguage2024}, and Faroese \citep{kunzFamilyMattersLanguage2026}. For Swedish, the remaining major Scandinavian language, CPT has been explored, but with limited methodological details \citep{aiSweden}, or for relatively small models \citep{glockerGrowMergeScaling2025, kunzPreferencesIdiomaticLanguage2026}. Instead, efforts have mainly targeted from-scratch pre-training \citep{norlundBuildingSwedishOpenDomain2021, ekgrenLessonsLearnedGPTSW32022, ekgrenGPTSW3AutoregressiveLanguage2024, kalpakchiSweCTRLMiniDatatransparentTransformerbased2023, amdVikingSailing}.

This paper aims to jointly address the understudied adaptation of LLMs to journalism and Swedish. By initially curating a high-quality corpus from millions of Swedish news articles, we can subsequently specialise models via CPT. Addressing the scarcity of evaluation resources, we introduce a domain-specific benchmark covering six editorial tasks to gauge the effectiveness of adaptation. Our results show that CPT increases models' generation capabilities and parametric knowledge, but hurts their performance on discriminative tasks. Moreover, we find the data-mixture composition to be crucial for successful adaptation, observing negative results when omitting experience replay. In our experiments, low-rank adaptation is competitive with full fine-tuning, and drastically more compatible with instruct vectors to improve instruction following. Notably, an existing Swedish benchmark fails to reflect the broadly positive in-domain effects of CPT, illustrating the need for targeted evaluation. Overall, we demonstrate that CPT is a viable approach for adapting LLMs to Swedish journalism, with the potential to enable publishers to develop and retain control over their editorial tools. We also provide insights and a blueprint for researchers and practitioners seeking to specialise models for other languages and domains.

\section{Background and Related Work}
\label{sec:background}
CPT is a canonical approach to enhancing a model's knowledge and capabilities in some target area \citep{gururanganDontStopPretraining2020}. A core challenge is to adapt a model without largely losing its original abilities, that is, to avoid catastrophic forgetting \citep{wuContinualLearningLarge2025}.

\subsection{Data Curation}
\label{sec:curation}
Experience replay \citep{chaudhryTinyEpisodicMemories2019} is commonly combined with CPT, mitigating forgetting by including a proportion as small as 1\% of previously seen examples during training \citep{scialomFinetunedLanguageModels2022}. Additionally, replaying data may even enhance target-domain performance \citep{ibrahimSimpleScalableStrategies2024, parmarReuseDontRetrain2024, keDemystifyingDomainadaptivePosttraining2025}. If a base model's pre-training distribution is unknown, replay data like code and English text are typically sampled from openly available corpora \citep{chenMEDITRON70BScalingMedical2023, etxanizLatxaOpenLanguage2024, dorkinEstLLMEnhancingEstonian2026}. The choice of such proxy data influences CPT efficacy, with higher-quality sources yielding larger in-domain gains \citep{wangLanguageAdaptationLarge2025}. Training on a well-curated subset rather than the full dataset can improve results further \citep{xieEfficientContinualPretraining2024, guoEfficientDomainContinual2025, nagEfficientContinualPretraining2025}. For pre-training in general, deduplicating data increases efficiency, reduces data leakage, causes less memorisation, and can ultimately boost performance \citep{leeDeduplicatingTrainingData2022, penedoFineWebDatasetsDecanting2024}.

\subsection{Constrained Learning}
\label{sec:constrained}
A complementary remedy for catastrophic forgetting is to restrain CPT to avoid excessive overwriting of knowledge \citep{guoComprehensiveSurveyContinual2025}. This regularising effect can be achieved by architecturally constraining parameter updates \citep{keAdaptingLanguageModel2022, keContinualPretrainingLanguage2023, nayakSculptingSubspacesConstrained2026} or carefully configuring learning rate dynamics \citep{guptaContinualPreTrainingLarge2023, parmarReuseDontRetrain2024}. Parameter-efficient fine-tuning (PEFT) methods such as low-rank adaptation (LoRA) \citep{huLoRALowRankAdaptation2022} may be used to this end \citep{nagEfficientContinualPretraining2025, pezeshkpourLearningSurfaceHow2025, kunzFamilyMattersLanguage2026}. Functionally restricting updates, LoRA shifts the balance of CPT to reduce forgetting at the expense of learning \citep{bidermanLoRALearnsLess2024, elhadyEmergentAbilitiesLarge2025}, sometimes to the extent of obstructing learning altogether \citep{tejaswiExploringDesignChoices2024}. Upscaling methods like LLaMA~Pro \citep{wuLLaMAProProgressive2024}, SOLAR \citep{kimSOLAR107BScaling2024}, and SCALE \citep{leeSCALEUpscaledContinual2026} serve a similar purpose by introducing parameters to expand a model's width or depth, while keeping its pre-trained backbone frozen during CPT.

\subsection{Instruction Following}
\label{sec:instruction}
Naively performing CPT can effectively erase a model's instruction-following capabilities \citep{jindalBalancingContinuousPreTraining2024}. To counteract this, a CPT corpus may be augmented with instruction-like or reading-comprehension data \citep{shiDontStopPretraining2023, chengInstructionPreTrainingLanguage2024, chengAdaptingLargeLanguage2024, rodriguezContinuedPretrainingInterpretabilityBased2025}. Given the scarcity of specialised instruction data, several training-free methods instead combine a CPT-adapted model with an instruction-tuned version of its base model, relying on either model merging \citep{siriwardhanaDomainAdaptationLlama370BInstruct2024, uedaMergingContinualPretraining2026} or model editing akin to task vectors \citep{ilharcoEditingModelsTask2023}. The latter subtracts the original pre-trained model from its instruction-tuned descendant in parameter space and adds the resulting instruct vector to the adapted model \citep{huangChatVectorSimple2024, linEfficientModelDevelopment2025, tanwarUnderstandingEffectsDomain2025}.

\subsection{Targeted Evaluation}
\label{sec:evaluation}
The effectiveness of CPT is commonly assessed using existing domain-specific benchmarks \citep{hiranoConstructionDomainSpecifiedJapanese2024, yingDataEfficientSelectionGrammatical2025, luoBioMedGPTOpenMultimodal2026}. For journalism, although general evaluation frameworks exist \citep{nishalDomainSpecificEvaluationStrategies2024}, previous work relies on either English financial benchmarks \citep{wuBloombergGPTLargeLanguage2023} or Chinese tasks \citep{wangMediaGPTLargeLanguage2023, liNewsBenchSystematicEvaluation2024, yaoNewsGPTLarge2024}. In Swedish, there exist merely three evaluation tasks related to journalism: lead paragraph generation \citep{monsenMethodBuildingNonEnglish2021}, as well as summarisation and search engine-optimised (SEO) headline generation \citep{eideSchibstedText2024}. For Swedish in general, the only existing benchmark collections are the Swedish portion of EuroEval \citep{nielsenScandEvalBenchmarkScandinavian2023} and Superlim \citep{berdicevskisSuperlimSwedishLanguage2023}, the latter primarily catered to encoder models.

\section{Method}
Drawing on work reviewed in Section~\ref{sec:background}, we create a domain-specific corpus and evaluation benchmark, and develop a blueprint for CPT adaptation.

\subsection{BonCorpus: Compiling Pre-Training Data}
\label{sec:boncorpus}
We source the pre-training corpus from 21.7M Swedish articles published by \href{https://www.bonniernews.se/home-en}{Bonnier News} publications from 1991 up until 2026. The publications include daily, evening, financial, and local newspapers, as well as lifestyle, sports, and trade magazines. Although individual texts are generally of high quality, the collection as a whole is noisy in several respects. For example, older articles originally stored in now-obsolete formats have been imperfectly migrated over time, leaving artefacts such as links in the plain-text body. Moreover, publications frequently cross-publish content, resulting in many near-duplicate articles. To alleviate these issues, we implement an extensive pre-processing pipeline. For a detailed description, see Appendix~\ref{sec:train-data}.

First, we parse and aggregate articles to standardise their format across publications. We then apply Unicode normalisation and resolve encoding inconsistencies before using regular expressions to remove extraneous text spans. To discard low-quality articles, we develop a broad set of filters based on text statistics, fastText \citep{joulinBagTricksEfficient2017} language-identification scores, and metadata such as tags and article types. Thereafter, we perform exact and fuzzy deduplication, using MinHash~LSH \citep{broderResemblanceContainmentDocuments1997} to identify candidate pairs for the latter. After calculating and thresholding the Levenshtein distance for each pair, we construct a graph with the retained pairs as edges and their constituent articles as nodes. We compute the graph's connected components to group similar articles and finally select only the longest one from each group. The full pipeline reduces the document count by nearly a third, yielding \textit{BonCorpus}: 14.6M articles totalling 8.5B tokens under the tokeniser of our base model family, which is specified in Section~\ref{sec:experiment}.

\subsection{BonEval: Constructing Evaluation Tasks}
\label{sec:boneval}
Due to the dearth of evaluation resources for Swedish journalism, we construct the \textit{BonEval} benchmark by repurposing editorial content into evaluation tasks. See Appendix~\ref{sec:eval-data} for more details on this process. BonEval comprises six tasks spanning three categories representative of editorial LLM application: generative tasks entailing text production, discriminative tasks targeting information extraction, and knowledge-intensive tasks involving factual accuracy. The six tasks are:

\begin{enumerate}[leftmargin=*, itemsep=0pt]
    \item \textbf{Headline:} headline generation based on the remainder of the article text.
    \item \textbf{Lead:} lead-paragraph generation based on the article body.
    \item \textbf{Summary:} bullet-point summarisation of the full article text.
    \item \textbf{Topic:} multi-class news topic classification of the full article text. The 19 labels extend the top-level terms in the Media Topics taxonomy \citepalias{iptcMediaTopics}.
    \item \textbf{Entity:} salient named entity recognition (NER) on the full article text. Unlike standard NER, only persons, organisations, and locations central to the article are targeted.
    \item \textbf{Quiz:} multiple-choice question answering based on general-knowledge quizzes. They span 15 categories and range from highly specific Swedish themes to globally known subjects.
\end{enumerate}

For all tasks except Quiz, we source evaluation examples from BonCorpus and subsequently exclude the corresponding articles from the training data. To minimise potential overlap with our base models' original pre-training data, while still covering major quadrennial events such as elections and sports championships, we only consider articles published from 2022 onward. For Headline, Lead, Topic, and Entities, we use stratified sampling to obtain 8,192 examples per task, approximately balanced across publications. Summary and Quiz each come from a single publication, and using all available data results in 100 and 11,960 examples, respectively.

\subsection{BonLM: Adapting Models}
\label{sec:bonlm}
To create the domain-specialised \textit{BonLM}, we perform CPT with BonCorpus as the primary training data. However, because the corpus exclusively comprises Swedish editorial content, it introduces a marked distribution shift relative to existing models’ pre-training data. To mitigate forgetting and potentially improve adaptation, we therefore employ experience replay to construct several data mixtures. Specifically, we sample code and English text from Common Corpus, a high-quality collection of uncopyrighted and openly licensed data \citep{langlaisCommonCorpusLargest2026}. Although editorial text in pre-training corpora can improve general language-modelling performance \citep{delarosaImpactCopyrightedMaterial2025}, it is unknown whether general-domain text can improve performance on editorial tasks. To investigate this, we also sample Swedish text from Common Corpus for our mixtures. Omitting the article-specific parsing and cleaning steps, we apply our pre-processing pipeline to all replay data.

We curate the data mixtures by diluting BonCorpus with every possible combination of the three replay types. Guided by prior work \citep{fujiiContinualPreTrainingCrossLingual2024, messmerEnhancingMultilingualLLM2025, samuelSmallLanguagesBig2025}, we sample each included type to constitute 10\% of the tokens in the resulting mixture. This yields eight mixtures of varying sizes, labelled by concatenating abbreviations for their constituents: BonCorpus (B), code (C), English (E), and Swedish (S). The full BonCorpus accounts for 100\% of the tokens in the smallest mixture, B, and 70\% in the largest, BCES.  We also study the impact of PEFT methods on CPT by comparing LoRA and LLaMA~Pro against full fine-tuning (FFT). Finally, we explore the effects of adding an instruct vector (IV) to each model.

\section{Experimental Setup}
\label{sec:experiment}
Following \citet{chenEffectiveEfficientContinual2025}, we conduct initial experiments with smaller models to derive a training recipe. As starting points for CPT, we use the 3B and 8B base models from Mistral's Ministral~3 family \citep{liuMinistral32026}. We defer details of our training setup to Appendix~\ref{sec:setup}.

\subsection{Evaluation}
We assess the effectiveness of CPT with BonEval. To validate our benchmark, we compare its results to those obtained on three existing tasks. These supplementary tasks are similar to ours but smaller in scale: article summarisation and SEO headline generation \citep{eideSchibstedText2024}, as well as question answering about Sweden-related facts \citep{kunzDiagnosticBenchmarkSwedenRelated2026}. Additionally, we evaluate models on seven Swedish tasks from EuroEval: SweReC for sentiment classification, SUC~3.0 for NER, ScaLA for linguistic acceptability, MultiWikiQA for reading comprehension, SweDN for summarisation, MMLU for knowledge, and HellaSwag for common-sense reasoning \citep{euroEvalSwedish}. Among these, SweDN partially overlaps with BonCorpus. For comparability, we implement BonEval following EuroEval's methodology, using few-shot prompting, multiple evaluation rounds, and the same primary metrics \citep{nielsenEncoderVsDecoder2025}. Specifically, we evaluate Headline, Lead, and Summary with \textsc{chrF3}++ \citep{popovicChrFWordsHelping2017}; Topic and Quiz with Matthews correlation coefficient (MCC); and Entity with micro-averaged F$_1$ score. See Appendix~\ref{sec:formulations} for implementation details.

\subsection{Initial Experiments}
For each of the eight data mixtures, we run CPT on Ministral~3~3B for 8,000 steps, approximating one full epoch over B; the other mixtures are larger and thus subsampled. Based on the average BonEval score across tasks, we select the best-performing mixture for subsequent experiments. Thereafter, we perform CPT with LoRA and LLaMA~Pro, configuring each method to approximately render equal numbers of trainable parameters. For the resulting models, we examine the effect of adding unweighted IVs. Since LLaMA~Pro alters the model architecture, there is no straightforward way to apply IV, and preliminary experimentation yielded poor outcomes. We also explored adding analogously computed reasoning vectors, but they generally underperformed their instruct counterparts.

\subsection{Final Training}

\begin{table}[!b]
\centering\small
\begin{tabular}{lc}
\toprule
\textbf{Model} & \textbf{Average} \\
\midrule
\href{https://hf.co/mistralai/Ministral-3-3B-Base-2512}{Ministral~3~3B Base} & 39.73 \\
\quad + BonCorpus (B) & 36.98 \\
\qquad + Code (BC) & 39.99 \\
\qquad + English (BE) & 40.01 \\
\qquad + Swedish (BS) & 40.06 \\
\qquad + Code + English (BCE) & 40.01 \\
\qquad + Code + Swedish (BCS) & 41.34 \\
\qquad + English + Swedish (BES) & 39.02 \\
\qquad + Code + English + Swedish (BCES) & \underline{41.48} \\
\midrule
LoRA & \underline{41.32} \\
LLaMA~Pro & 41.17 \\
\midrule
Full fine-tuning (FFT) + instruct vector (IV) & 39.07 \\
LoRA + IV & \underline{\textbf{42.27}} \\
\bottomrule
\end{tabular}
\caption{\label{tab:initial}Average BonEval scores from 3B experiments. The best mixture (BCES) is used for the results in the last four rows. \underline{Underline} denotes section-best and \textbf{bold} overall best. See full results in Table~\ref{tab:initial-boneval}, Appendix~\ref{sec:eval-all}.}
\end{table}

\begin{table*}[t]
\centering\small
\begin{tabular}{lcccccccc}
\toprule
\textbf{Model} & \textbf{SweReC} & \textbf{SUC~3.0} & \textbf{ScaLA} & \textbf{MultiWikiQA} & \textbf{SweDN} & \textbf{MMLU} & \textbf{HellaSwag} & \textbf{Average} \\
\midrule
\href{https://hf.co/AI-Sweden-Models/gpt-sw3-6.7b-v2}{GPT-SW3~6.7B} & 10.44 & 26.72 & 9.84 & 69.47 & 30.62 & 2.92 & 2.05 & 21.72 \\
\href{https://hf.co/AI-Sweden-Models/Llama-3-8B}{Llama~SW3~8B} & \underline{79.90} & 36.02 & 8.81 & 72.22 & 32.23 & 17.02 & 5.64 & 35.98 \\
\href{https://hf.co/swiss-ai/Apertus-8B-2509}{Apertus~8B} & 79.11 & \underline{45.74} & \underline{34.34} & \underline{74.05} & \underline{33.39} & \underline{41.54} & \underline{32.54} & \underline{48.67} \\
\midrule
\multicolumn{9}{l}{\textit{Ministral~3~8B}} \\
\midrule
\href{https://hf.co/mistralai/Ministral-3-8B-Base-2512}{Base} & \underline{80.19} & \underline{66.44} & \underline{51.70} & 75.71 & 35.30 & \underline{\textbf{58.10}} & 43.68 & 58.73 \\
\href{https://hf.co/mistralai/Ministral-3-8B-Instruct-2512}{Instruct} & 77.53 & 60.51 & 50.62 & 74.28 & \underline{35.31} & 56.22 & 57.27 & 58.82 \\
\href{https://hf.co/mistralai/Ministral-3-8B-Reasoning-2512}{Reasoning} & 77.05 & 65.21 & 47.77 & \underline{\textbf{79.44}} & 31.16 & 49.96 & \underline{\textbf{64.27}} & \underline{\textbf{59.26}} \\
\midrule
\multicolumn{9}{l}{\textit{BonLM~8B}} \\
\midrule
FFT & 80.01 & 50.64 & 39.61 & \underline{70.53} & 31.94 & 30.62 & 17.66 & 45.86 \\
FFT + IV & 79.07 & 45.84 & 22.54 & 70.04 & 36.03 & 20.09 & 12.48 & 40.87 \\
LoRA & \textbf{\underline{80.55}} & \textbf{\underline{66.86}} & \textbf{\underline{59.85}} & 69.25 & 35.03 & \underline{53.34} & 36.82 & \underline{57.39} \\
LoRA + IV & 78.27 & 62.27 & 54.24 & 70.05 & \textbf{\underline{36.27}} & 47.18 & \underline{52.06} & 57.19 \\
\bottomrule
\end{tabular}
\caption{\label{tab:euroeval}Results for BonLM~8B and baselines on Swedish EuroEval tasks. Micro-averaged F$_1$ is reported for SUC~3.0, token-averaged F$_1$ for MultiWikiQA, and \textsc{chrF3}++ for SweDN. For all other tasks, MCC is reported. \underline{Underline} denotes section-best and \textbf{bold} overall best.
}
\end{table*}

To obtain BonLM~8B, we run CPT on Ministral~3~8B using the best recipe from the initial experiments. We set the number of training steps to the equivalent of four epochs, a duration shown to be effective for CPT in prior work \citep{muennighoffScalingDataConstrainedLanguage2023, xueRepeatNotRepeat2023, guoEfficientDomainContinual2025}. Thereafter, we evaluate the resulting model on all tasks and compare it with similarly sized baselines: the base, instruct, and reasoning versions of Ministral~3~8B; GPT-SW3~6.7B, pre-trained from scratch on Scandinavian-language data \citep{ekgrenGPTSW3AutoregressiveLanguage2024}; Llama~SW3~8B, adapted to the Scandinavian languages via CPT \citep{aiSweden}; and Apertus~8B, pre-trained on multilingual data \citep{hernandez-canoApertusDemocratizingOpen2026}.

\section{Results and Analysis}
The main experimental results are presented in Tables~\ref{tab:initial},~\ref{tab:euroeval},~and~\ref{tab:boneval}, with additional details provided in Appendix~\ref{sec:eval-all}.

\subsection{Impact of Data-Mixture Composition}
Table~\ref{tab:initial} reports average BonEval scores from the initial 3B experiments. Overall, CPT improves upon the base model for all mixtures except B and BES. B exhibits the greatest degradation, likely reflecting catastrophic forgetting caused by the mixture's homogeneous, BonCorpus-only composition. Conversely, BCES includes all text types and achieves the highest score, narrowly outperforming BCS. These observations corroborate the findings reviewed in Section~\ref{sec:curation} that replaying code and English text enhances target-domain performance. Furthermore, the results suggest that replaying general-domain text benefits performance on editorial tasks. The task-level breakdown in Table~\ref{tab:initial-boneval}, Appendix~\ref{sec:eval-all}, show that adding Swedish text to any mixture consistently raises scores on Lead and especially Quiz. We surmise that the gain on Quiz arises because Common Corpus includes knowledge-dense sources such as Swedish Wikipedia, which cover detailed facts that complement those typically found in news articles. For the remaining experiments, we perform CPT with the BCES mixture.

\subsection{PEFT and Instruct-Vector Compatibility}
Among the PEFT methods in Table~\ref{tab:initial}, LoRA slightly outperforms LLaMA~Pro despite having fewer trainable parameters, as detailed in Appendix~\ref{sec:peft-config}. Although this result diverges from the findings of \citet{wuLLaMAProProgressive2024}, later work has reported better target-domain adaptation with LoRA than with LLaMA~Pro \citep{leeSCALEUpscaledContinual2026}. Adding an IV to the LoRA model further improves performance, yielding the highest average BonEval score. By contrast, adding an IV to the FFT model impairs performance, contrary to prior work described in Section~\ref{sec:instruction}. To examine LoRA's regularising effect, we compute the mean relative Euclidean distance between the adapted and base-model parameters, obtaining 0.08 for LoRA and 0.18 for FFT. We hypothesise that this smaller drift partly explains LoRA's compatibility with IV, as the resulting model remains closer in parameter space to the model that originally underwent instruction tuning. Why IV is counterproductive for the FFT model remains unclear. Because prior studies have focused primarily on Llama and Qwen models, IV compatibility may vary across model families, but further investigation is needed to confirm this. Given that FFT outperforms PEFT without IV, and LoRA with IV performs best overall, we train one BonLM~8B variant using each method.

\begin{table*}[!t]
\centering\small
\begin{tabular}{lcccccccccc}
\toprule
\multirow{2.6}{*}{\textbf{Model}} & \multicolumn{7}{c}{\textbf{BonEval}} & \multicolumn{3}{c}{\textbf{Supplementary}} \\
\cmidrule(r){2-8} \cmidrule(l){9-11}
& \textbf{Headline} & \textbf{Lead} & \textbf{Summary} & \textbf{Topic} & \textbf{Entity} & \textbf{Quiz} & \textbf{Average} & \textbf{SvD} & \textbf{AB} & \textbf{Facts} \\
\midrule
\href{https://hf.co/AI-Sweden-Models/gpt-sw3-6.7b-v2}{GPT-SW3~6.7B} & 17.70 & 22.78 & 36.11 & 0.49 & \underline{27.50} & 6.00 & 18.43 & 15.96 & 33.87 & 1.69 \\
\href{https://hf.co/AI-Sweden-Models/Llama-3-8B}{Llama~SW3~8B} & \underline{19.35} & \underline{29.13} & 40.51 & 68.28 & 26.17 & \underline{61.30} & \underline{40.79} & 18.57 & 36.17 & \underline{23.54} \\
\href{https://hf.co/swiss-ai/Apertus-8B-2509}{Apertus~8B} & 18.61 & 23.09 & \underline{40.93} & \underline{70.95} & 22.81 & 61.05 & 39.57 & \underline{21.09} & \underline{40.07} & 18.38 \\
\midrule
\multicolumn{11}{l}{\textit{Ministral~3~8B}} \\
\midrule
\href{https://hf.co/mistralai/Ministral-3-8B-Base-2512}{Base} & 18.35 & 25.79 & 42.49 & \textbf{\underline{72.84}} & \textbf{\underline{29.97}} & \underline{60.12} & 41.59 & 20.08 & 41.88 & 20.62 \\
\href{https://hf.co/mistralai/Ministral-3-8B-Instruct-2512}{Instruct} & \underline{29.28} & \underline{32.03} & 46.27 & 71.93 & 23.20 & 57.76 & \underline{43.41} & \textbf{\underline{30.41}} & \underline{45.10} & \underline{25.21} \\
\href{https://hf.co/mistralai/Ministral-3-8B-Reasoning-2512}{Reasoning} & 21.36 & 27.68 & \underline{47.93} & 71.06 & 20.67 & 52.91 & 40.27 & 11.86 & 42.86 & 19.27 \\
\midrule
\multicolumn{11}{l}{\textit{BonLM~8B}} \\
\midrule
FFT & 21.58 & 29.88 & 43.76 & \underline{72.40} & 28.34 & 68.86 & 44.14 & 21.81 & 39.80 & 40.82 \\
FFT + IV & 24.32 & 31.24 & 43.26 & 42.28 & 28.38 & 61.98 & 38.58 & 22.79 & 40.29 & 32.41 \\
LoRA & 21.29 & 31.07 & 44.35 & 68.93 & \underline{29.81} & \textbf{\underline{71.36}} & 44.47 & 23.85 & 43.60 & \textbf{\underline{42.24}} \\
LoRA + IV & \textbf{\underline{29.76}} & \textbf{\underline{35.01}} & \textbf{\underline{49.88}} & 70.54 & 23.99 & 66.95 & \textbf{\underline{46.02}} & \underline{30.14} & \textbf{\underline{47.18}} & 38.56 \\
\bottomrule
\end{tabular}
\caption{\label{tab:boneval}Results for BonLM~8B and baselines on BonEval and supplementary tasks Svenska Dagbladet (\textbf{SvD}) SEO Headline, Aftonbladet (\textbf{AB}) Summary, and Swedish\textbf{Facts}. Micro-averaged F$_1$ is reported for Entity, and MCC for Topic, Quiz, and SwedishFacts. For all other tasks, \textsc{chrF3}++ is reported.}
\end{table*}

\subsection{Mixed Results on EuroEval}
The results for BonLM~8B and the baselines on the Swedish EuroEval tasks are shown in Table~\ref{tab:euroeval}. Ministral Base outperforms the three other baselines across all tasks. It also surpasses its post-trained descendants on all tasks except reading comprehension, summarisation, and common-sense reasoning. On average, both BonLMs underperform their base model, plausibly due to CPT-induced catastrophic forgetting. Similarly, LoRA's substantial outperformance of FFT may stem from the method's regularising effect described in Section~\ref{sec:constrained}. For individual tasks, CPT causes drastic regressions on the LLM-translated MMLU and HellaSwag, and a more modest decline on the LLM-synthesised MultiWikiQA. Contrastively, BonLM LoRA narrowly beats the base model on the remaining tasks, all of which are natively Swedish.

\subsection{Improved Performance on BonEval}
Table~\ref{tab:boneval} displays the results on BonEval and the supplementary tasks for BonLM~8B and the baselines. Again, Ministral Base generally outperforms the other baselines, although Llama~SW3 and Apertus achieve higher scores on Headline and Quiz. BonLM FFT and LoRA both outperform all baselines on average and on most individual tasks. Unlike in the initial 3B experiments and in prior work \citep{bidermanLoRALearnsLess2024, tejaswiExploringDesignChoices2024}, LoRA is generally superior to FFT. The effect of IV mirrors that observed for the 3B models, and LoRA~+~IV yields the highest average score. At the task level, CPT consistently improves the generative Headline, Lead, and Summary but slightly degrades the discriminative Topic and Entity. We conjecture that classification and NER are comparatively domain-agnostic, as there may be no uniquely journalistic way to perform them. If so, retaining the base model's general capabilities likely benefits performance on these tasks. As with the 3B models, Quiz exhibits the largest absolute gains from CPT. This suggests that news articles contain culturally rich information and that LLMs can be imbued with such knowledge through CPT.

The supplementary-task results support the validity of BonEval, with CPT improving the base model's generative abilities and factual knowledge. A minor discrepancy is Ministral Instruct's strong performance on SEO headline generation. More broadly, Table~\ref{tab:boneval} reveals a consistent trend: instruction tuning increases generation quality. Across all five generative tasks, Instruct outperforms Base, and LoRA~+~IV surpasses LoRA. The modest results for Reasoning are expected since BonEval does not target reasoning capabilities. Nevertheless, prolonged post-training of Ministral notably reduces scores on Topic, Entity, and Quiz.

\subsection{Value of Targeted Evaluation}
In summary, CPT improves overall performance on BonEval but degrades it on EuroEval. We partly attribute this disparity to BonLM acquiring domain-specific capabilities at the expense of general ones. For example, CPT on predominantly editorial content unsurprisingly decreases the models' common-sense reasoning performance, as measured by HellaSwag. However, machine-translated benchmarks can exhibit cultural biases \citep{singhGlobalMMLUUnderstanding2025}, overestimate target-language performance \citep{kuulmetsTranslatedBenchmarksCan2023}, and miss nuances captured by native evaluations \citep{chenItGoodData2024}. This raises the possibility that the LLM-synthesised portion of EuroEval does not accurately reflect a model's proficiency in Swedish. On knowledge tasks specifically, CPT drives substantial gains on the native Quiz and SwedishFacts but causes regressions on the translated MMLU. Importantly, EuroEval does not fully capture the in-domain benefits of CPT evident in BonEval, such as stronger generative capabilities and greater factual knowledge. This underscores the importance of language- and domain-specific benchmarks for guiding model adaptation.

\section{Conclusion and Future Work}
This paper presents the first comprehensive study of adapting LLMs to Swedish journalism. By detailing the curation of BonCorpus, which spans 35 years of editorial content, we highlight the challenges involved in processing ostensibly clean articles. We further introduce BonEval, a domain-specific benchmark comprising six editorial tasks. Our results reaffirm that CPT is effective for domain adaptation, but only when coupled with carefully constructed experience-replay mixtures. Across two model sizes, CPT consistently improves performance on generative and knowledge-intensive tasks but not on discriminative tasks. We also find that IV compatibility depends on the adaptation method: IV generally benefits models trained with LoRA but not those trained with FFT. Crucially, our findings demonstrate the importance of integrating targeted evaluation into the adaptation process, as EuroEval's Swedish tasks fail to reflect the broader in-domain trends revealed by BonEval.

We conclude that journalism is a promising domain for LLM research and application. In future work, we plan to investigate the downstream impact of CPT by fine-tuning BonLM for editorial use cases such as headline generation and stylistic correction. We also recognise the limitations of automatic evaluation and aggregate metrics. Going forward, we intend to collaborate with journalists to design human-evaluation protocols that better assess the real-world utility of LLMs in newsrooms.

\section*{Limitations}
For business and legal reasons, we cannot release proprietary data, code, or models, which reduces the reproducibility of our study. Computational constraints also restrict our adaptation experiments to a single model family, limiting the generalisability of our findings. Because Ministral already has relatively strong Swedish-language capabilities, experiments with another model family could yield different conclusions. Moreover, because the original pre-training data are undisclosed, we cannot assess potential overlap with BonCorpus or BonEval, nor its effects on our results. We also provide limited insight into the downstream effects of CPT since we do not fine-tune BonLM for specific editorial applications. Finally, our exclusive reliance on automatic evaluation precludes any assertive claims about real-world utility.

\section*{Ethical Considerations}
Because our research artefacts will not be released, the general risk of misuse by external actors is limited. However, BonLM is intended for internal deployment as part of Bonnier News' editorial tools. It would be subject to strict usage policies but nonetheless affect the work of reporters and editors and, ultimately, the news presented to the public. Although adapting existing LLMs is more efficient than training them from scratch, it still requires substantial computational resources, as detailed in Appendix~\ref{sec:setup}. Yet we view a strong base model as an investment that can improve computational efficiency in future work. Finally, copyright is an existential concern for publishers, and we have carefully considered it throughout the data-sourcing process.

\section*{Acknowledgments}
This work was partially supported by the Wallenberg AI, Autonomous Systems and Software Program (WASP) funded by the Knut and Alice Wallenberg Foundation (KAW), and by TrustLLM funded by Horizon Europe GA 101135671. Computational resources were provided on the Berzelius system funded by KAW and operated by the National Academic Infrastructure for Supercomputing in Sweden (NAISS). We are grateful to Hans Hjelm at Bonnier News for his support and guidance throughout the project.

\bibliography{custom}

\appendix

\section{Training Data}
\label{sec:train-data}
This section describes selected stages of the pre-processing pipeline in greater detail.

\subsection{Cleaning}
We clean the articles by omitting extraneous fields during source-format parsing and by applying regular-expression-based filters. In particular, we remove text spans that are peripheral to the main article content, such as links, advertisements, reader notices, and author bylines. Additionally, we remove artefacts erroneously retained as plain text, such as JavaScript code and HTML tags. By contrast, we preserve elements that provide additional context, including fact boxes, tables, and lists.

\subsection{Filtering}
To remove low-quality or otherwise unsuitable articles, we apply a broad set of text- and metadata-based filters to the raw corpus. We design these filters using a combination of domain knowledge and exploratory analysis. When articles of a known category cannot be identified from metadata alone, we instead develop text-based statistics and regular expressions to detect them. By analysing articles flagged as candidates for removal, we iteratively tune the filters to balance false positives and false negatives. Additionally, by examining articles with extreme values for various textual properties, such as unusually high newline counts, we identify and remove previously unknown categories of low-quality content. For example, some general text-quality filters require each article to
\begin{itemize}[itemsep=0pt]
\item have a headline;
\item have body text;
\item contain between 2 and 2,000 newline characters, inclusive, counting those following the headline or lead paragraph;
\item contain at least 20 words;
\item contain at least 17 distinct words;
\item contain at least 200 characters;
\item have a type--token ratio of at least 0.2; and
\item be classified as Swedish by fastText\footnote{\url{https://hf.co/facebook/fasttext-language-identification}} with a score greater than 0.9.
\end{itemize}

Although articles are sourced from Swedish publications, some are written in other languages to serve specific readerships; using the fastText filter, we remove content in languages such as English, Arabic, and Russian. The largest category of excluded material consists of template-based articles that are automatically generated from structured data, typically covering sports results, real estate transactions, and weather forecasts. In addition, we discard large numbers of roundup articles consisting primarily of links to previously published content on a particular topic. Other article types we remove include republished press releases and public records, as well as personal notices such as birthday greetings, wedding announcements, and obituaries.

\subsection{Deduplication}
We perform document-level deduplication on normalised text obtained by lowercasing and removing punctuation and diacritics. Following \citet{penedoFineWebDatasetsDecanting2024}, we construct MinHash signatures using 112 hash functions and apply LSH for candidate generation, but lower the similarity threshold to 0.25 to increase recall. For each candidate pair, we compute the normalised Levenshtein distance and treat the articles as duplicates if it is less than 0.2.

\section{Evaluation Data}
\label{sec:eval-data}
When constructing BonEval, we apply task-specific filters to obtain suitable and representative evaluation examples. Across all tasks sampled from BonCorpus, we exclude articles longer than 1,024 tokens. For each task, we apply the following additional criteria:

\begin{enumerate}[align=left, leftmargin=*, itemsep=0pt]
\item \textbf{Headline}: Headlines contain 10--120 characters, and input texts contain more than 200 characters. We exclude headlines beginning with ``LIST: \dots'' or similar prefixes.

\item \textbf{Lead}: Leads contain 30--500 characters, and input texts contain more than 200 characters. Each lead must also be shorter than half its corresponding body text. We exclude leads containing bullet points of any kind.

\item \textbf{Summary}: Summaries contain 200--800 characters and comprise three to five bullet points. We normalise their formatting.

\item \textbf{Topic}: Articles are originally labelled with one to five topics, exactly one of which is top-level and serves as the target. To simplify the underlying hierarchical multi-label classification problem, which contains thousands of topics, we restrict the label set to the 19 top-level topics shown in Table~\ref{tab:base-prompts-en}.

\item \textbf{Entity}: Articles have one to five entity labels, each of which is mentioned within the first 90\% of the text. We exclude labels likely to reflect annotation errors, such as external news outlets derived from attribution phrases like ``according to \dots''. Additionally, for geographically nested location labels, we retain only the more specific one; for example, we keep Stockholm rather than Sweden when both are present.

\item \textbf{Quiz}: Each question has exactly four answer choices. We discard quizzes with an English-language theme or that rely on images or audio. Because the quizzes date back to 2009, we also attempt to exclude strongly time-dependent content, including quizzes in the category ``Current affairs'' and questions whose answers change over time, such as ``How many prime ministers has the United Kingdom had?''. The quizzes span the following 15 categories:

\begin{enumerate}[align=left, leftmargin=*, itemsep=0pt]
\item Words and language
\item Current affairs (removed)
\item Around Sweden
\item Around the world
\item Animals and nature
\item Science
\item Food and drink
\item Sport and games
\item Books
\item Film and TV
\item Music
\item Culture
\item Politics and society
\item History
\item Miscellaneous
\end{enumerate}
\end{enumerate}

\begin{table}[t]
\centering\small
\begin{tabular}{ll}
\toprule
\textbf{Hyperparameter} & \textbf{Value} \\ 
\midrule
$\beta_1$ & 0.9 \\
$\beta_2$ & 0.95 \\
$\epsilon$ & 1e-8 \\
Weight decay & 0.1\\
Schedule & Cosine \\
Linear warmup & 10\% \\
Max. learning rate & 5e-5 \\
Min. learning rate & 5e-6 \\
Dropout & 0.0 \\ 
Gradient clipping & 1.0 \\
Sequence length & 16,384 \\
Global batch size & 64 \\
Precision & bfloat16 \\
\bottomrule
\end{tabular}
\caption{\label{tab:hyperparameters}Training configuration for all experiments.}
\end{table}

\section{Training Setup}
\label{sec:setup}
Across all experiments, we use NVIDIA DGX H200 nodes and consume a total of 6,500 GPU-hours. We use the AdamW optimiser \citep{loshchilovDecoupledWeightDecay2019} and the hyperparameters reported in Table~\ref{tab:hyperparameters}. To improve computational and memory efficiency, we use DeepSpeed ZeRO-2 \citep{rajbhandariZeROMemoryOptimizations2020}, FlashAttention-3 \citep{shahFlashAttention3FastAccurate2024}, and Liger kernels \citep{hsuLigerKernelEfficientTriton2025}. Depending on the model size and number of GPUs, we use activation checkpointing and gradient accumulation as needed to achieve the target global batch size. Because article lengths vary substantially, we pack examples up to the maximum sequence length so that each optimisation step is based on roughly the same number of tokens. To minimise truncation, which can degrade model performance, we use a best-fit-decreasing packing strategy \citep{dingFewerTruncationsImprove2024}. As the packed sequences still vary slightly in length, we flatten each batch into a single sequence to eliminate padding inefficiencies \citep{kunduEnhancingTrainingEfficiency2024}.

\subsection{PEFT Configurations}
\label{sec:peft-config}
We configure LoRA and LLaMA~Pro to have comparable numbers of trainable parameters for the 3B model: 395M and 466M, respectively. For LLaMA~Pro, we insert one additional layer after every sixth layer of the base model. Following \citet{bidermanLoRALearnsLess2024}, we apply LoRA to all transformer modules with a rank of 256, an $\alpha$ of 512, and no dropout.

\section{Evaluation Setup}
\label{sec:formulations}
We implement BonEval and run the evaluations with the \href{https://euroeval.com/}{EuroEval} library. For tasks outside BonEval, we rely on EuroEval's existing implementations. To parse model outputs, we use log probabilities for Topic and Quiz and structured generation for Entity.

We evaluate each model over 10 runs, randomly sampling few-shot examples from the training split for each run. During development, we evaluate on validation splits rather than test splits. Following \citet{nielsenEncoderVsDecoder2025}, we heuristically determine the number of few-shot examples for each task based on context length. Rather than targeting 1,000 tokens, we select the number of examples such that no input exceeds 4,096 tokens. We cap this number at 12, guided by the existing task implementations.\footnote{\url{https://github.com/EuroEval/EuroEval/blob/e630b62/src/euroeval/tasks.py}} The resulting number of few-shot examples for each task is reported in Table~\ref{tab:boneval-details}. If IV is applied to a model, we consider it instruction-tuned. Task prompts for base and instruction-tuned models are shown in Tables~\ref{tab:base-prompts} and~\ref{tab:instruct-prompts}, respectively; English translations are provided in Tables~\ref{tab:base-prompts-en} and~\ref{tab:instruct-prompts-en}, respectively. For Quiz, we use EuroEval's default multiple-choice prompts.

\begin{table}[t]
\centering\small
\begin{tabular}{lllll}
\toprule
\textbf{Task} & \textbf{Train} & \textbf{Val} & \textbf{Test} & \textbf{Few-Shot} \\
\midrule
Headline & 1,024 & 256 & 8,192 & 3 \\
Lead & 1,024 & 256 & 8,192 & 3 \\
Summary & 32 & 32 & 100 & 3 \\
Topic & 1,024 & 256 & 8,192 & 3 \\
Entity & 1,024 & 256 & 8,192 & 3 \\
Quiz & 1,024 & 256 & 11,960 & 12 \\
\bottomrule
\end{tabular}
\caption{\label{tab:boneval-details}Number of examples for BonEval tasks.}
\end{table}

\begin{table*}
\centering\small
\begin{tabular}{lll}
\toprule
\textbf{Task} & \textbf{Prompt Prefix} & \textbf{Prompt Template}\\
\midrule
Headline & \makecell[lt]{Följande är artiklar med\\tillhörande rubriker.} & \makecell[lt]{Artikel: \texttt{\{lead + body\}} \\ Rubrik: \texttt{\{headline\}}} \\
\midrule
Lead & \makecell[lt]{Följande är artiklar med\\tillhörande ingresser.} & \makecell[lt]{Artikel: \texttt{\{body\}} \\ Ingress: \texttt{\{lead\}}} \\
\midrule
Summary & \makecell[lt]{Följande är artiklar med\\tillhörande sammanfattningar\\i punktform.} & \makecell[lt]{Artikel: \texttt{\{article\}} \\ Sammanfattning: \texttt{\{summary\}}} \\
\midrule
Topic & \makecell[lt]{Följande är artiklar med\\tillhörande frågor och svar\\om deras ämnen.} & \makecell[lt]{Artikel: \texttt{\{article\}}\\Fråga: Vilket ämne beskriver bäst artikelns innehåll?\\Svarsalternativ:\\\makecell[l]{%
    \begin{tabular}{@{}l@{\enspace}l@{}}
        a. & Arbete \\
        b. & Brott, lag och rätt \\
        c. & Ekonomi, näringsliv och finans \\
        d. & Hälsa och sjukvård \\
        e. & Klimat och miljö \\
        f. & Konflikter och krig \\
        g. & Kultur och nöje \\
        h. & Kungligt \\
        i. & Livsstil och fritid \\
        j. & Olyckor och katastrofer \\
        k. & Personligt \\
        l. & Politik \\
        m. & Religion och tro \\
        n. & Samhälle \\
        o. & Skola och utbildning \\
        p. & Sport \\
        q. & Trafik och fordon \\
        r. & Vetenskap och teknik \\
        s. & Väder
    \end{tabular}%
}\\Svar:\\\texttt{\{a|b|c|d|e|f|g|h|i|j|k|l|m|n|o|p|q|r|s\}}} \\
\midrule
Entity & \makecell[lt]{Följande är artiklar med\\tillhörande namngivna\\entiteter i JSON-format\\som är centrala för\\innehållet.} & \makecell[lt]{Artikel: \texttt{\{article\}} \\ Namngivna entiteter: \texttt{\{}\\\texttt{\quad"person": [...],}\\\texttt{\quad"plats": [...],}\\\texttt{\quad"organisation": [...]}\\\texttt{\}}} \\
\midrule
Quiz & \makecell[lt]{Följande är flervalsfrågor\\(med svar).} & \makecell[lt]{Fråga: \texttt{\{question\}}\\Svarsalternativ:\\a. \texttt{\{choice\_a\}}\\b. \texttt{\{choice\_b\}}\\c. \texttt{\{choice\_c\}}\\d. \texttt{\{choice\_d\}}\\Svar: \texttt{\{a|b|c|d\}}} \\
\bottomrule
\end{tabular}
\caption{\label{tab:base-prompts}BonEval prompts for base models.}
\end{table*}

\begin{table*}
\centering\small
\begin{tabular}{ll}
\toprule
\textbf{Task} & \textbf{Instruction Prompt}\\
\midrule
Headline & \makecell[lt]{Artikel: \texttt{\{lead + body\}}\\Skriv en rubrik till artikeln ovan.} \\
\midrule
Lead & \makecell[lt]{Artikel: \texttt{\{body\}}\\Skriv en ingress till artikeln ovan.} \\
\midrule
Summary & \makecell[lt]{Artikel: \texttt{\{article\}}\\Skriv en sammanfattning i punktform av artikeln ovan.} \\
\midrule
Topic & \makecell[lt]{Artikel: \texttt{\{article\}}\\Fråga: Vilket ämne beskriver bäst artikelns innehåll?\\Svarsalternativ:\\\makecell[l]{%
    \begin{tabular}{@{}l@{\enspace}l@{}}
        a. & Arbete \\
        b. & Brott, lag och rätt \\
        c. & Ekonomi, näringsliv och finans \\
        d. & Hälsa och sjukvård \\
        e. & Klimat och miljö \\
        f. & Konflikter och krig \\
        g. & Kultur och nöje \\
        h. & Kungligt \\
        i. & Livsstil och fritid \\
        j. & Olyckor och katastrofer \\
        k. & Personligt \\
        l. & Politik \\
        m. & Religion och tro \\
        n. & Samhälle \\
        o. & Skola och utbildning \\
        p. & Sport \\
        q. & Trafik och fordon \\
        r. & Vetenskap och teknik \\
        s. & Väder
    \end{tabular}%
}\\Besvara frågan ovan med\\a, b, c, d, e, f, g, h, i, j, k, l, m, n, o, p, q, r eller s,\\och inget annat.} \\
\midrule
Entity & \makecell[lt]{Artikel: \texttt{\{article\}}\\Identifiera de namngivna entiteter som är centrala för\\artikelns innehåll. Svara i JSON-format med nycklarna\\person, plats och organisation, där värdena är listor över\\de namngivna entiterna av den typen, precis som de\\förekommer i artikeln.} \\
\midrule
Quiz & \makecell[lt]{Fråga: \texttt{\{question\}}\\Besvara frågan ovan med a, b, c eller d,\\och inget annat.} \\
\bottomrule
\end{tabular}
\caption{\label{tab:instruct-prompts}BonEval prompts for instruction-tuned models.}
\end{table*}

\begin{table*}
\centering\small
\begin{tabular}{lll}
\toprule
\textbf{Task} & \textbf{Prompt Prefix} & \textbf{Prompt Template}\\
\midrule
Headline & \makecell[lt]{The following are articles\\with associated headlines.} & \makecell[lt]{Article: \texttt{\{lead + body\}} \\ Headline: \texttt{\{headline\}}} \\
\midrule
Lead & \makecell[lt]{The following are articles\\with associated leads.} & \makecell[lt]{Article: \texttt{\{body\}} \\ Lead: \texttt{\{lead\}}} \\
\midrule
Summary & \makecell[lt]{The following are articles\\with associated bullet-point\\summaries.} & \makecell[lt]{Article: \texttt{\{article\}}\\Summary: \texttt{\{summary\}}} \\
\midrule
Topic & \makecell[lt]{The following are articles\\with associated questions\\ and answers about\\their topics.} & \makecell[lt]{Article: \texttt{\{article\}}\\Question: Which topic best describes\\the content of the article?\\Choices:\\\makecell[l]{%
    \begin{tabular}{@{}l@{\enspace}l@{}}
        a. & Labour \\
        b. & Crime, law and justice \\
        c. & Economy, business and finance \\
        d. & Health \\
        e. & Environment \\
        f. & Conflict, war and peace \\
        g. & Arts, culture, entertainment and media \\
        h. & Royalty \\
        i. & Lifestyle and leisure \\
        j. & Disaster, accident and emergency incident \\
        k. & Human interest \\
        l. & Politics and government \\
        m. & Religion \\
        n. & Society \\
        o. & Education \\
        p. & Sport \\
        q. & Traffic and vehicles \\
        r. & Science and technology \\
        s. & Weather
    \end{tabular}%
}\\Answer:\\\texttt{\{a|b|c|d|e|f|g|h|i|j|k|l|m|n|o|p|q|r|s\}}} \\
\midrule
Entity & \makecell[lt]{The following are articles\\with associated named\\entities in JSON format\\that are central to the\\content.} & \makecell[lt]{Article: \texttt{\{article\}} \\ Named entities: \texttt{\{}\\\texttt{\quad"person": [...],}\\\texttt{\quad"location": [...],}\\\texttt{\quad"organisation": [...]}\\\texttt{\}}} \\
\midrule
Quiz & \makecell[lt]{The following are multiple\\choice questions\\(with answers).} & \makecell[lt]{Question: \texttt{\{question\}}\\Choices:\\a. \texttt{\{choice\_a\}}\\b. \texttt{\{choice\_b\}}\\c. \texttt{\{choice\_c\}}\\d. \texttt{\{choice\_d\}}\\Answer: \texttt{\{a|b|c|d\}}} \\
\bottomrule
\end{tabular}
\caption{\label{tab:base-prompts-en}Translated BonEval prompts for base models.}
\end{table*}

\begin{table*}
\centering\small
\begin{tabular}{ll}
\toprule
\textbf{Task} & \textbf{Instruction Prompt}\\
\midrule
Headline & \makecell[lt]{Article: \texttt{\{lead + body\}}\\Write a headline to the above article.} \\
\midrule
Lead & \makecell[lt]{Article: \texttt{\{body\}}\\Write a lead to the above article.} \\
\midrule
Summary & \makecell[lt]{Article: \texttt{\{article\}}\\Write a bullet-point summary of the above article.} \\
\midrule
Topic & \makecell[lt]{Article: \texttt{\{article\}}\\Question: Which topic best describes\\the content of the article?\\Choices:\\\makecell[l]{%
    \begin{tabular}{@{}l@{\enspace}l@{}}
        a. & Labour \\
        b. & Crime, law and justice \\
        c. & Economy, business and finance \\
        d. & Health \\
        e. & Environment \\
        f. & Conflict, war and peace \\
        g. & Arts, culture, entertainment and media \\
        h. & Royalty \\
        i. & Lifestyle and leisure \\
        j. & Disaster, accident and emergency incident \\
        k. & Human interest \\
        l. & Politics and government \\
        m. & Religion \\
        n. & Society \\
        o. & Education \\
        p. & Sport \\
        q. & Traffic and vehicles \\
        r. & Science and technology \\
        s. & Weather
    \end{tabular}%
}\\Answer the above question by replying with\\a, b, c, d, e, f, g, h, i, j, k, l, m, n, o, p, q, r, or s,\\and nothing else.} \\
\midrule
Entity & \makecell[lt]{Article: \texttt{\{article\}}\\Identify the named entities central to the content of\\the article. Respond in JSON format with the keys\\person, location, and organisation, where the values\\are lists of the named entities of that type, exactly\\as they appear in the article.} \\
\midrule
Quiz & \makecell[lt]{Question: \texttt{\{question\}}\\Answer the above question by replying with a, b, c or d,\\and nothing else.} \\
\bottomrule
\end{tabular}
\caption{\label{tab:instruct-prompts-en}Translated BonEval prompts for instruction-tuned models.}
\end{table*}

\section{Complete Evaluation Results}
\label{sec:eval-all}
Complete evaluation results for the 3B experiments are presented in Tables~\ref{tab:initial-euroeval} and~\ref{tab:initial-boneval}, while those for BonLM~8B and the baselines are presented in Tables~\ref{tab:euroeval-all} and~\ref{tab:boneval-all}.

\begin{table*}[t]
\centering\small
\begin{tabular}{lcccccccc}
\toprule
\textbf{Model} & \textbf{SweReC} & \textbf{SUC~3.0} & \textbf{ScaLA} & \textbf{MultiWikiQA} & \textbf{SweDN} & \textbf{MMLU} & \textbf{HellaSwag} & \textbf{Average} \\
\midrule
\multicolumn{9}{l}{\textit{Ministral~3~3B}} \\
\midrule
\href{https://hf.co/mistralai/Ministral-3-3B-Base-2512}{Base} & \underline{77.61} & \textbf{\underline{55.79}} & \textbf{\underline{38.86}} & 75.41 & \underline{34.70} & \textbf{\underline{49.04}} & 26.02 & \textbf{\underline{51.06}} \\
\href{https://hf.co/mistralai/Ministral-3-3B-Instruct-2512}{Instruct} & 75.23 & 39.66 & 33.50 & 70.94 & 32.66 & 42.32 & \textbf{\underline{48.18}} & 48.93 \\
\href{https://hf.co/mistralai/Ministral-3-3B-Reasoning-2512}{Reasoning} & 74.17 & 44.48 & 29.58 & \textbf{\underline{78.11}} & 22.62 & 26.10 & 35.57 & 44.37 \\
\midrule
\multicolumn{9}{l}{\textit{Data Mixtures}} \\
\midrule
B & 78.22 & 31.41 & 28.22 & 65.83 & 30.77 & 21.24 & 5.80 & 37.35 \\
BC & 75.09 & 41.52 & 28.30 & 71.09 & 32.34 & 24.70 & 10.03 & 40.44 \\
BE & 75.68 & 39.46 & 22.84 & 72.28 & 33.12 & 24.07 & 8.21 & 39.38 \\
BCE & 78.62 & 47.04 & 25.50 & \underline{72.41} & 33.58 & 26.78 & 7.20 & 41.59 \\
BS & 77.30 & 41.92 & 25.04 & 69.16 & \underline{35.11} & 26.31 & \underline{13.80} & 41.24 \\
BCS & \textbf{\underline{79.60}} & 47.24 & \underline{35.09} & 68.33 & 34.94 & 27.31 & 11.74 & \underline{43.46} \\
BES & 79.25 & 45.77 & 33.25 & 69.28 & 34.99 & 28.61 & 10.26 & 43.06 \\
BCES & 78.85 & \underline{47.60} & 32.31 & 70.43 & 33.46 & \underline{28.90} & 11.49 & 43.29 \\
\midrule
\multicolumn{9}{l}{\textit{PEFT Methods}} \\
\midrule
LoRA & 77.39 & \underline{55.46} & 33.50 & \underline{75.18} & 34.37 & 44.46 & 20.30 & 48.67 \\
LLaMA~Pro & \underline{78.04} & 54.17 & \underline{34.09} & 74.73 & \underline{34.48} & \underline{47.22} & \underline{20.92} & \underline{49.09} \\
\midrule
\multicolumn{9}{l}{\textit{Instruct Vectors}} \\
\midrule
B + IV & 75.23 & 27.13 & 7.33 & 54.50 & 27.85 & 7.58 & 4.21 & 29.12 \\
BC + IV & 77.10 & 36.87 & 9.77 & 72.08 & 32.24 & 10.97 & 6.24 & 35.04 \\
BE + IV & 74.11 & 33.89 & 10.23 & 72.28 & 30.79 & 9.69 & 4.28 & 33.61 \\
BCE + IV & 77.63 & \underline{50.36} & 12.82 & 69.90 & 33.24 & 16.57 & 9.44 & 38.57 \\
BS + IV & 77.83 & 44.04 & 15.27 & 71.25 & 33.74 & 13.42 & 9.40 & 37.85 \\
BCS + IV & 74.58 & 42.88 & 14.63 & 72.20 & \textbf{\underline{35.87}} & 13.26 & 6.58 & 37.14 \\
BES + IV & 69.12 & 40.72 & 26.07 & 69.68 & 32.18 & 17.40 & 7.82 & 37.57 \\
BCES + IV & 75.71 & 44.60 & 23.91 & \underline{72.87} & 34.03 & 17.49 & 9.38 & 39.71 \\
LoRA + IV & \underline{78.31} & 46.00 & \underline{32.76} & 71.57 & 34.79 & \underline{37.71} & \underline{34.86} & \underline{48.00} \\
\bottomrule
\end{tabular}
\caption{\label{tab:initial-euroeval}Results from 3B experiments on Swedish EuroEval tasks. Micro-averaged F$_1$ is reported for SUC~3.0, token-averaged F$_1$ for MultiWikiQA, and \textsc{chrF3}++ for SweDN. For all other tasks, MCC is reported. \underline{Underline} denotes section-best and \textbf{bold} overall best.}
\end{table*}

\begin{table*}
\centering\small
\begin{tabular}{lcccccccccc}
\toprule
\multirow{2.6}{*}{\textbf{Model}} & \multicolumn{7}{c}{\textbf{BonEval}} & \multicolumn{3}{c}{\textbf{Supplementary}} \\
\cmidrule(r){2-8} \cmidrule(l){9-11}
& \textbf{Headline} & \textbf{Lead} & \textbf{Summary} & \textbf{Topic} & \textbf{Entity} & \textbf{Quiz} & \textbf{Average} & \textbf{SvD} & \textbf{AB} & \textbf{Facts} \\
\midrule
\multicolumn{8}{l}{\textit{Ministral~3~3B}} \\
\midrule
\href{https://hf.co/mistralai/Ministral-3-8B-Base-2512}{Base} & 18.62 & 26.24 & 42.28 & 70.39 & \underline{29.80} & \underline{51.07} & 39.73 & 19.94 & 40.72 & 15.81 \\
\href{https://hf.co/mistralai/Ministral-3-8B-Instruct-2512}{Instruct} & \underline{\textbf{26.38}} & 31.28 & 45.87 & \underline{71.80} & 23.15 & 47.80 & \underline{41.05} & \underline{27.96} & \underline{44.15} & \underline{16.64} \\
\href{https://hf.co/mistralai/Ministral-3-8B-Reasoning-2512}{Reasoning} & 21.19 & \underline{28.80} & \underline{46.08} & 68.45 & 23.49 & 33.66 & 36.94 & 14.70 & 39.59 & 10.15 \\
\midrule
\multicolumn{8}{l}{\textit{Data Mixtures}} \\
\midrule
B & 20.61 & 25.37 & 39.42 & 52.49 & 23.68 & 60.30 & 36.98 & 17.56 & 34.15 & 31.25 \\
BC & 20.48 & 22.63 & 40.83 & 67.61 & 28.40 & 59.97 & 39.99 & 20.02 & 37.92 & 30.61 \\
BE & 20.54 & 25.03 & 41.04 & 67.59 & 26.38 & 59.46 & 40.01 & 18.97 & 38.26 & 30.10 \\
BS & 19.81 & 27.28 & 42.87 & 62.87 & 26.11 & \underline{\textbf{61.41}} & 40.06 & 19.98 & 38.20 & 31.65 \\
BCE & 20.48 & 25.86 & 40.77 & 65.33 & 29.05 & 58.58 & 40.01 & 18.92 & 40.19 & 32.02 \\
BCS & \underline{20.98} & 26.74 & \underline{43.42} & 68.35 & 27.65 & 60.89 & 41.34 & 19.76 & 40.21 & \textbf{\underline{33.74}} \\
BES & 20.26 & \underline{27.63} & 40.80 & 60.92 & 23.73 & 60.80 & 39.02 & 17.73 & \underline{40.30} & 30.78 \\
BCES & 20.73 & 26.83 & 42.24 & \underline{70.44} & \underline{29.28} & 59.40 & \underline{41.48} & \underline{20.46} & 39.48 & 32.93 \\
\midrule
\multicolumn{8}{l}{\textit{PEFT Methods}} \\
\midrule
LoRA & 19.95 & \underline{27.24} & \underline{43.98} & 69.32 & \underline{\textbf{30.47}} & \underline{56.97} & \underline{41.32} & \underline{20.95} & \underline{41.12} & \underline{24.74} \\
LLaMA~Pro & \underline{20.24} & 26.47 & 43.05 & \underline{\textbf{71.85}} & 30.30 & 55.09 & 41.17 & 19.97 & 40.10 & 23.57 \\
\midrule
\multicolumn{8}{l}{\textit{Instruct Vectors}} \\
\midrule
B + IV & 23.15 & 28.41 & 39.57 & 1.48 & 23.66 & 43.48 & 26.62 & 17.89 & 29.10 & 14.56 \\
BC + IV & 25.61 & 29.06 & 41.34 & 47.30 & 25.66 & 48.36 & 36.22 & 25.74 & 38.25 & 18.26 \\
BE + IV & 25.24 & 32.62 & 40.86 & 42.60 & 21.61 & 44.31 & 34.54 & 24.85 & 39.14 & 18.27 \\
BCE + IV & 24.99 & \underline{\textbf{32.76}} & 43.20 & 46.74 & 25.42 & 49.19 & 37.05 & 19.87 & 40.91 & 25.88 \\
BS + IV & 25.38 & 30.79 & 41.77 & 7.64 & 23.93 & \underline{54.35} & 30.64 & 20.88 & 35.57 & 25.97 \\
BCS + IV & 24.00 & 31.79 & 44.56 & 16.38 & 23.34 & 49.76 & 31.64 & 23.92 & 41.66 & 22.28 \\
BES + IV & 23.48 & 31.34 & 41.06 & 31.18 & 21.38 & 51.23 & 33.28 & 20.35 & 38.32 & 24.38 \\
BCES + IV & 25.09 & 31.86 & 44.72 & 55.34 & \underline{26.06} & 51.36 & 39.07 & 20.08 & 40.75 & \underline{27.66} \\
LoRA + IV & \underline{26.06} & 31.55 & \underline{\textbf{48.20}} & \underline{69.33} & 25.10 & 53.36 & \underline{\textbf{42.27}} & \textbf{\underline{30.14}} & \textbf{\underline{45.60}} & 22.45 \\
\bottomrule
\end{tabular}
\caption{\label{tab:initial-boneval}Results from 3B experiments on BonEval and supplementary tasks Svenska Dagbladet (\textbf{SvD}) SEO
Headline, Aftonbladet (\textbf{AB}) Summary, and Swedish\textbf{Facts}. Micro-averaged F$_1$ is reported for Entity, and MCC for
Topic, Quiz, and SwedishFacts. For all other tasks, \textsc{chrF3}++ is reported. \underline{Underline} denotes section-best and \textbf{bold} overall best.}
\end{table*}

\begin{table*}
\centering\small
\begin{tabular}{lccc}
\toprule
\textbf{Model} & \textbf{SweReC} &\textbf{SUC~3.0} & \textbf{ScaLA} \\
\midrule
\href{https://hf.co/AI-Sweden-Models/gpt-sw3-6.7b-v2}{GPT-SW3~6.7B} & 10.44 ± 5.52 / 13.17 ± 4.04 & 26.72 ± 2.58 / 20.29 ± 2.65 & 9.84 ± 1.72 / 50.83 ± 3.01 \\
\href{https://hf.co/AI-Sweden-Models/Llama-3-8B}{Llama~SW3~8B} &79.90 ± 0.86 / 75.36 ± 1.95 & 36.02 ± 2.16 / 26.67 ± 2.22 & 8.81 ± 2.26 / 53.34 ± 0.97 \\
\href{https://hf.co/swiss-ai/Apertus-8B-2509}{Apertus~8B} & 79.11 ± 0.42 / 75.56 ± 1.40 & 45.74 ± 2.02 / 31.99 ± 2.86 & 34.34 ± 4.28 / 63.41 ± 4.61 \\
\midrule
\multicolumn{4}{l}{\textit{Ministral~3~8B}} \\
\midrule
\href{https://hf.co/mistralai/Ministral-3-8B-Base-2512}{Base} & 80.19 ± 0.89 / 77.61 ± 1.68 & 66.44 ± 2.73 / 56.13 ± 5.14 & 51.70 ± 3.18 / 72.75 ± 2.57 \\
\href{https://hf.co/mistralai/Ministral-3-8B-Instruct-2512}{Instruct} & 77.53 ± 1.40 / 77.72 ± 1.60 & 60.51 ± 1.80 / 37.49 ± 3.12 & 50.62 ± 1.62 / 73.58 ± 1.53 \\
\href{https://hf.co/mistralai/Ministral-3-8B-Reasoning-2512}{Reasoning} & 77.05 ± 1.23 / 78.58 ± 0.96 & 65.21 ± 2.35 / 46.84 ± 5.92 & 47.77 ± 2.11 / 73.38 ± 1.13 \\
\midrule
\multicolumn{4}{l}{\textit{BonLM~8B}} \\
\midrule
FFT & 80.01 ± 1.19 / 78.68 ± 0.83 & 50.64 ± 2.54 / 39.69 ± 4.46 & 39.61 ± 3.35 / 68.07 ± 2.49 \\
FFT + IV & 79.07 ± 1.51 / 79.96 ± 0.90 & 45.84 ± 3.33 / 30.85 ± 3.45 & 22.54 ± 2.08 / 52.49 ± 3.10 \\
LoRA & 80.55 ± 0.71 / 79.68 ± 1.10 & 66.86 ± 2.40 / 57.70 ± 4.66 & 59.85 ± 2.35 / 79.35 ± 1.45 \\
LoRA + IV & 78.27 ± 0.99 / 79.74 ± 0.72 & 62.27 ± 1.73 / 37.58 ± 3.26 & 54.24 ± 2.12 / 76.68 ± 1.16 \\
\bottomrule
\newline
\end{tabular}

\begin{tabular}{lcc}
\toprule
\textbf{Model} & \textbf{MMLU} & \textbf{HellaSwag} \\
\midrule
\href{https://hf.co/AI-Sweden-Models/gpt-sw3-6.7b-v2}{GPT-SW3~6.7B} & 2.92 ± 0.78 / 24.38 ± 0.74 & 2.05 ± 1.26 / 25.91 ± 0.80 \\
\href{https://hf.co/AI-Sweden-Models/Llama-3-8B}{Llama~SW3~8B} &17.02 ± 1.24 / 37.27 ± 0.87 & 5.64 ± 1.44 / 29.15 ± 1.02 \\
\href{https://hf.co/swiss-ai/Apertus-8B-2509}{Apertus~8B} & 41.54 ± 1.04 / 55.81 ± 0.85 & 32.54 ± 1.80 / 48.61 ± 1.52 \\
\midrule
\multicolumn{3}{l}{\textit{Ministral~3~8B}} \\
\midrule
\href{https://hf.co/mistralai/Ministral-3-8B-Base-2512}{Base} & 58.10 ± 0.90 / 68.44 ± 0.67 & 43.68 ± 3.19 / 56.16 ± 2.74 \\
\href{https://hf.co/mistralai/Ministral-3-8B-Instruct-2512}{Instruct} & 56.22 ± 1.32 / 66.93 ± 1.01 & 57.27 ± 1.22 / 67.88 ± 0.92 \\
\href{https://hf.co/mistralai/Ministral-3-8B-Reasoning-2512}{Reasoning} & 49.96 ± 1.61 / 62.39 ± 1.20 & 64.27 ± 1.57 / 73.00 ± 1.27 \\
\midrule
\multicolumn{3}{l}{\textit{BonLM~8B}} \\
\midrule
FFT & 30.62 ± 0.94 / 47.93 ± 0.74 & 17.66 ± 2.17 / 37.31 ± 1.66 \\
FFT + IV & 20.09 ± 2.11 / 39.49 ± 2.08 & 12.48 ± 3.46 / 32.92 ± 3.22 \\
LoRA & 53.34 ± 0.91 / 64.80 ± 0.63 & 36.82 ± 2.96 / 50.75 ± 2.62 \\
LoRA + IV & 47.18 ± 1.50 / 59.86 ± 1.16 & 52.06 ± 1.60 / 63.80 ± 1.20 \\
\bottomrule
\newline
\end{tabular}

\begin{tabular}{lcc}
\toprule
\textbf{Model} & \textbf{MultiWikiQA} & \textbf{SweDN} \\
\midrule
\href{https://hf.co/AI-Sweden-Models/gpt-sw3-6.7b-v2}{GPT-SW3~6.7B} & 69.47 ± 1.95 / 54.28 ± 2.52 & 30.62 ± 1.09 / 32.89 ± 1.26 \\
\href{https://hf.co/AI-Sweden-Models/Llama-3-8B}{Llama~SW3~8B} &72.22 ± 1.74 / 56.16 ± 2.15 & 32.23 ± 0.86 / 33.14 ± 1.03 \\
\href{https://hf.co/swiss-ai/Apertus-8B-2509}{Apertus~8B} & 74.05 ± 1.99 / 55.93 ± 2.66 & 33.39 ± 0.85 / 35.03 ± 0.98 \\
\midrule
\multicolumn{3}{l}{\textit{Ministral~3~8B}} \\
\midrule
\href{https://hf.co/mistralai/Ministral-3-8B-Base-2512}{Base} & 75.71 ± 2.89 / 59.71 ± 3.45 & 35.30 ± 1.13 / 37.30 ± 1.45 \\
\href{https://hf.co/mistralai/Ministral-3-8B-Instruct-2512}{Instruct} & 74.28 ± 1.75 / 56.32 ± 2.07 & 35.31 ± 0.20 / 38.54 ± 0.20 \\
\href{https://hf.co/mistralai/Ministral-3-8B-Reasoning-2512}{Reasoning} & 79.44 ± 1.93 / 64.01 ± 2.54 & 31.16 ± 1.28 / 37.39 ± 1.12 \\
\midrule
\multicolumn{3}{l}{\textit{BonLM~8B}} \\
\midrule
FFT & 70.53 ± 1.40 / 52.92 ± 1.31 & 31.94 ± 1.85 / 32.89 ± 2.11 \\
FFT + IV & 70.04 ± 1.89 / 56.36 ± 2.21 & 36.03 ± 0.83 / 38.20 ± 1.05 \\
LoRA & 69.25 ± 2.27 / 51.67 ± 2.68 & 35.03 ± 1.14 / 36.84 ± 1.31 \\
LoRA + IV & 70.05 ± 2.09 / 50.23 ± 2.61 & 36.27 ± 0.33 / 39.57 ± 0.36 \\
\bottomrule
\end{tabular}
\caption{\label{tab:euroeval-all}Results for BonLM~8B and baselines on Swedish EuroEval tasks, showing primary and secondary metrics with 95\% confidence intervals. MCC / macro-averaged F$_1$ are reported for SweRec and ScaLA, micro-averaged F$_1$ without / with MISC for SUC~3.0, MCC / accuracy for MMLU and HellaSwag, token-averaged F$_1$ / exact match for MultiWikiQA, and \textsc{chrF3}++ / \textsc{chrF4}++ for SweDN.}
\end{table*}

\begin{table*}
\centering\small
\begin{tabular}{lccc}
\toprule
\textbf{Model} & \textbf{Headline} & \textbf{Lead} & \textbf{Summary} \\
\midrule
\href{https://hf.co/AI-Sweden-Models/gpt-sw3-6.7b-v2}{GPT-SW3~6.7B} & 17.70 ± 0.98 / 17.85 ± 1.06 & 22.78 ± 1.79 / 23.36 ± 2.02 & 36.11 ± 1.93 / 37.17 ± 2.13 \\
\href{https://hf.co/AI-Sweden-Models/Llama-3-8B}{Llama~SW3~8B} &19.35 ± 0.97 / 19.41 ± 1.01 & 29.13 ± 1.33 / 29.86 ± 1.39 & 40.51 ± 2.08 / 40.90 ± 2.30 \\
\href{https://hf.co/swiss-ai/Apertus-8B-2509}{Apertus~8B} & 18.61 ± 0.88 / 18.66 ± 0.92 & 23.09 ± 2.42 / 23.22 ± 2.59 & 40.93 ± 2.23 / 41.60 ± 2.45 \\
\midrule
\multicolumn{4}{l}{\textit{Ministral~3~8B}} \\
\midrule
\href{https://hf.co/mistralai/Ministral-3-8B-Base-2512}{Base} & 18.35 ± 0.74 / 18.40 ± 0.77 & 25.79 ± 2.22 / 26.07 ± 2.41 & 42.49 ± 3.06 / 43.01 ± 3.44 \\
\href{https://hf.co/mistralai/Ministral-3-8B-Instruct-2512}{Instruct} & 29.28 ± 0.21 / 31.02 ± 0.29 & 32.03 ± 0.80 / 33.19 ± 1.04 & 46.27 ± 0.45 / 48.74 ± 0.43 \\
\href{https://hf.co/mistralai/Ministral-3-8B-Reasoning-2512}{Reasoning} & 21.36 ± 1.50 / 21.58 ± 1.58 & 27.68 ± 1.71 / 27.92 ± 1.88 & 47.93 ± 1.39 / 49.31 ± 1.68 \\
\midrule
\multicolumn{4}{l}{\textit{BonLM~8B}} \\
\midrule
FFT & 21.58 ± 1.25 / 21.69 ± 1.30 & 29.88 ± 2.55 / 30.44 ± 2.75 & 43.76 ± 3.09 / 44.20 ± 3.36 \\
FFT + IV & 24.32 ± 1.68 / 24.56 ± 1.77 & 31.24 ± 1.49 / 31.71 ± 1.69 & 43.26 ± 2.30 / 43.53 ± 2.49 \\
LoRA & 21.29 ± 0.91 / 21.38 ± 0.94 & 31.07 ± 1.90 / 32.01 ± 2.14 & 44.35 ± 2.94 / 45.03 ± 3.30 \\
LoRA + IV & 29.76 ± 0.56 / 30.73 ± 0.66 & 35.01 ± 0.38 / 36.79 ± 0.62 & 49.88 ± 0.84 / 51.32 ± 1.05 \\
\bottomrule
\\
\end{tabular}

\begin{tabular}{lccc}
\toprule
\textbf{Model} & \textbf{Topic} & \textbf{Entity} & \textbf{Quiz} \\
\midrule
\href{https://hf.co/AI-Sweden-Models/gpt-sw3-6.7b-v2}{GPT-SW3~6.7B} & 0.49 ± 0.20 / 0.28 ± 0.06 & 27.50 ± 1.45 / 27.50 ± 1.45 & 6.00 ± 1.11 / 28.72 ± 1.03 \\
\href{https://hf.co/AI-Sweden-Models/Llama-3-8B}{Llama~SW3~8B} &68.28 ± 0.76 / 47.73 ± 0.78 & 26.17 ± 0.64 / 26.17 ± 0.64 & 61.30 ± 0.69 / 70.83 ± 0.54 \\
\href{https://hf.co/swiss-ai/Apertus-8B-2509}{Apertus~8B} & 70.95 ± 0.61 / 51.75 ± 0.84 & 22.81 ± 2.58 / 22.81 ± 2.58 & 61.05 ± 0.62 / 70.74 ± 0.49 \\
\midrule
\multicolumn{4}{l}{\textit{Ministral~3~8B}} \\
\midrule
\href{https://hf.co/mistralai/Ministral-3-8B-Base-2512}{Base} & 72.84 ± 0.76 / 59.84 ± 1.36 & 29.97 ± 0.96 / 29.97 ± 0.96 & 60.12 ± 0.56 / 69.96 ± 0.43 \\
\href{https://hf.co/mistralai/Ministral-3-8B-Instruct-2512}{Instruct} & 71.93 ± 1.02 / 59.39 ± 1.25 & 23.20 ± 0.28 / 23.20 ± 0.28 & 57.76 ± 0.55 / 68.08 ± 0.43 \\
\href{https://hf.co/mistralai/Ministral-3-8B-Reasoning-2512}{Reasoning} & 71.06 ± 1.17 / 55.46 ± 1.20 & 20.67 ± 1.24 / 20.67 ± 1.24 & 52.91 ± 0.43 / 64.50 ± 0.33 \\
\midrule
\multicolumn{4}{l}{\textit{BonLM~8B}} \\
\midrule
FFT & 72.40 ± 0.58 / 53.92 ± 1.28 & 28.34 ± 1.07 / 28.34 ± 1.07 & 68.86 ± 0.33 / 76.59 ± 0.25 \\
FFT + IV & 42.28 ± 9.30 / 29.82 ± 6.42 & 28.38 ± 0.82 / 28.38 ± 0.82 & 61.98 ± 0.62 / 71.34 ± 0.51 \\
LoRA & 68.93 ± 0.77 / 57.49 ± 0.95 & 29.81 ± 0.57 / 29.81 ± 0.57 & 71.36 ± 0.35 / 78.49 ± 0.27 \\
LoRA + IV & 70.54 ± 1.53 / 58.21 ± 1.48 & 23.99 ± 0.27 / 23.99 ± 0.27 & 66.95 ± 0.61 / 75.09 ± 0.47 \\
\bottomrule
\end{tabular}

\caption{\label{tab:boneval-all}Results for BonLM~8B and baselines on BonEval, showing primary and secondary metrics with 95\% confidence intervals. Micro-averaged F$_1$ without / with MISC are reported for Entity, MCC / macro-averaged F$_1$ for Topic, and MCC / accuracy for Quiz. For all other tasks, \textsc{chrF3}++ / \textsc{chrF4}++ is reported.}
\end{table*}

\end{document}